%% file: main.tex
\documentclass[letterpaper, 10 pt, conference]{ieeeconf}  

\IEEEoverridecommandlockouts                              

\usepackage{graphics} 
\usepackage{graphicx}
\usepackage{epsfig} 
\usepackage{times} 
\usepackage{amsmath} 
\usepackage{amssymb}  

\usepackage{bm}
\usepackage[linesnumbered,ruled,vlined]{algorithm2e}
\usepackage[noend]{algpseudocode}
\usepackage[font={small}]{caption}
\usepackage{subcaption}
\usepackage{physics}
\usepackage{xcolor}
\usepackage{tikz}
\usetikzlibrary {arrows.meta}
\usetikzlibrary{fit,
                positioning}
\usepackage[belowskip=1ex]{subcaption}  

\usetikzlibrary{calc} 
\usetikzlibrary{positioning}
\usepackage{balance}
\usepackage{soul}
\usetikzlibrary{arrows}
\usepackage{cite}
\usepackage{url}
\usepackage{multirow}
\usepackage{booktabs} 
\usepackage[hidelinks]{hyperref}

\input{tex/preamble.tex}

\usepackage{setspace}
\title{\LARGE \bf

Direction Informed Trees (DIT*): Optimal Path Planning via \\Direction Filter and Direction Cost Heuristic 
}

\author{Liding Zhang$^{1}$, Kejia Chen$^{1}$, Kuanqi Cai$^{1}$, Yu Zhang$^{1}$, Yixuan Dang$^{1}$, Yansong Wu$^{1}$, \\Zhenshan Bing$^{1,2}$, Fan Wu$^{1}$, Sami Haddadin$^{1\ddagger}$, Alois Knoll$^{1\dagger}$ 
\thanks{$^{1}$L. Zhang, K. Chen, K. Cai, Y. Zhang, Y. Dang, Y. Wu, Z. Bing, F. Wu,  S. Haddadin and A. Knoll are with the Department of Informatics, Technical University of Munich, Germany.
{\tt\small liding.zhang@tum.de}}%
\thanks{$^{2}$Z. Bing is also with the State Key Laboratory for Novel Software Technology and the School of Science and Technology, Nanjing University (Suzhou Campus), China.~\textit{(Corresponding author: Zhenshan Bing.)}}%
}
\begin{document}

\maketitle
\thispagestyle{empty}
\pagestyle{empty}

\begin{abstract}
Optimal path planning requires finding a series of feasible states from the starting point to the goal to optimize objectives. Popular path planning algorithms, such as Effort Informed Trees (EIT*), employ effort heuristics to guide the search. Effective heuristics are accurate and computationally efficient, but achieving both can be challenging due to their conflicting nature. This paper proposes Direction Informed Trees (DIT*), a sampling-based planner that focuses on optimizing the search direction for each edge, resulting in goal bias during exploration. We define edges as generalized vectors and integrate similarity indexes to establish a directional filter that selects the nearest neighbors and estimates direction costs. The estimated direction cost heuristics are utilized in edge evaluation. This strategy allows the exploration to share directional information efficiently. DIT* convergence faster than existing single-query, sampling-based planners on tested problems in $\mathbb{R}^4$ to $\mathbb{R}^{16}$ and has been demonstrated in real-world environments with various planning tasks. A video showcasing our experimental results is available at: \href{https://youtu.be/2SX6QT2NOek}{\textcolor{blue}{https://youtu.be/2SX6QT2NOek}}.
\\
%
\vspace{-0.5em}
\end{abstract}

\input{sections/sec1_intro}

\input{sections/sec2_1_related}
\input{sections/sec2_2_statement}
\input{sections/sec3_formulation}
\input{sections/sec4_analysis}

\input{sections/sec5_experiment}
\input{sections/sec6_conclusion}

\input{sections/sec7_acknowledgements}





\bibliographystyle{IEEEtran}
\bibliography{references}

\end{document}

%% file: tex/preamble.tex
\usepackage{amssymb}
\usepackage{amsfonts}
\usepackage{amsmath}
\usepackage{amsthm}
\usepackage{bm}

\input{tex/comments.tex}

\input{tex/symbols}

%% file: tex/comments.tex
\usepackage[normalem]{ulem}                                        
\usepackage{marginnote}
\usepackage[textwidth=10ex,colorinlistoftodos]{todonotes}

\colorlet{mh}{red}
\colorlet{fwu}{red}
\colorlet{ywu}{blue}
\colorlet{kchen}{blue}
\colorlet{lchen}{green}
\colorlet{zbing}{green}
\colorlet{shaddadin}{purple}
\colorlet{iperez}{cyan}
\colorlet{schneider}{magenta}



%% file: tex/symbols.tex
\usepackage{amssymb}
\usepackage{mathtools}
\usepackage{sansmath}

\mathchardef\mhyphen="2D   

\newcommand{\RNum}[1]{\uppercase\expandafter{\romannumeral #1\relax}}

%% file: sections/sec1_intro.tex
\section{Introduction}
Robot motion planning aims to find collision-free paths from start to goal configurations while avoiding obstacles~\cite{gammell2021asymptotically}. Search-based planners, such as A*~\cite{hart1968formal}, can't directly apply in high-dimensional spaces due to the \textit{curse of dimensionality}~\cite{bellman1957dynamic}. To address this challenge, sampling-based planners like Probabilistic Roadmaps (PRM)~\cite{kavraki1998analysis} and Rapidly-exploring Random Trees (RRT)~\cite{LaValle1998} generate feasible paths by randomly sampling vertices in the \textit{configuration space} (\textit{$\mathcal{C}$-space}) and connecting them using collision-checked local planners. However, sampling-based methods can require significant time to converge to an optimal solution. In practical applications, it is crucial to rapidly find a high-quality path to the goal configuration in order to conserve limited power.

\subsection{Related Work}
Recently, many researchers have proposed effective heuristics to accelerate path convergence.
RRT-Connect~\cite{kuffner2000rrt} search with bidirectional trees, using connection heuristics for faster path finding~\cite{zhang25g3t}.
Cost heuristics are applied in Heuristically-guided RRT (hRRT)~\cite{urmson2003approaches} and Generalized Bidirectional RRT (GBRRT)~\cite{nayak2020bidirectional}. hRRT uses prior heuristics to guide exploration within RRT's Voronoi regions, while GBRRT employs reverse tree-computed heuristics to guide the forward tree. Despite improved performance, these algorithms do not guarantee solution quality bounds~\cite{strub2022adaptively}.
RRT*~\cite{karaman2011sampling} improves on RRT by incrementally rewiring the tree to achieve asymptotic optimality. Informed RRT*~\cite{gammell2014informed} refines this by using elliptical informed sampling~\cite{gammell2018informed}, thus accelerating optimal path convergence. It adapts for kinodynamic systems~\cite{kunz2016hierarchical, yi2018generalizing}, nearest neighbor rewiring~\cite{Zhang2024Elliptical}, {greedy domain subsets}~\cite{Phone2022greedyinformed} and non-parametric informations\cite{joshi2019non}.
\begin{figure}[t]
    \centering
    \begin{tikzpicture}

    \node[inner sep=0pt] (russell) at (0,0)
    {\includegraphics[width=0.49\textwidth]{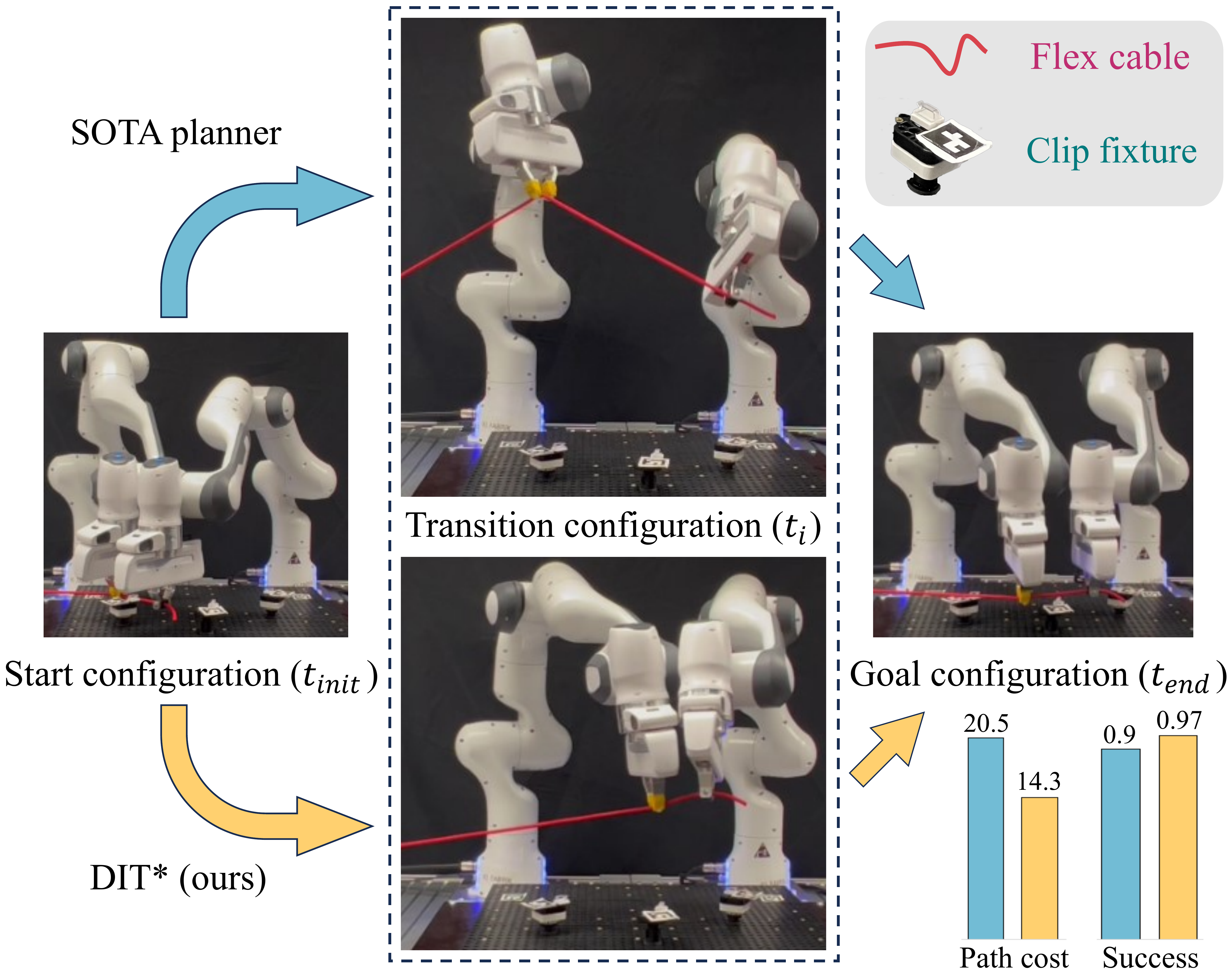}};


    
    \end{tikzpicture}
    \caption{Planner comparison on dual-arm manipulator robot during a real-time cable routing task in fixture clipping board scenarios.}
    \label{fig: dual_arm_setup}
    \vspace{-1.7em} 
\end{figure}
Batch Informed Trees (BIT*)~\cite{gammell2015batch, gammell2020batch} conceptualizes states as dense edge-implicit random geometric graphs (RGGs)~\cite{penrose2003random} and employs an admissible cost heuristic for efficient graph search. Advanced BIT*~\cite{strub2020advanced} enhances initial solution convergence by inflating its heuristics. Flexible Informed Trees (FIT*)~\cite{Zhang2024adaptive} uses adaptive batch sizes for rapid initial solution.
Adaptively Informed Trees (AIT*)~\cite{strub2022adaptively, strub2020adaptively} utilizes bidirectional asymmetric search with lazy-reverse search to accelerate exploration through information provided to the forward search.
Effort Informed Trees (EIT*)~\cite{strub2022adaptively} introduces admissible cost (i.e., the lower bound on the actual value) and effort heuristics during reverse search, currently standing as state-of-the-art (SOTA). However, these SOTA planners lack the ability to select an optimal search direction during the planning process. Optimizing path search directions is advisable, given the variations among knowledge dimensions~\cite{meng2025preserving,zhang2025TASE}, planning scenarios~\cite{Bing2023TPAMI,Bing2023TIE}, and bio-inspired robots~\cite{Bing2023SR,long2025robot}. This limitation hinders overall planning efficiency~\cite{zhang2024review}.

\subsection{Proposed Algorithm and Original Contribution}
\begin{figure}[t!]
    \centering
    \begin{tikzpicture}
    
    \node[inner sep=0pt] (russell) at (0.0,0.0)
    {\includegraphics[width=0.48\textwidth]{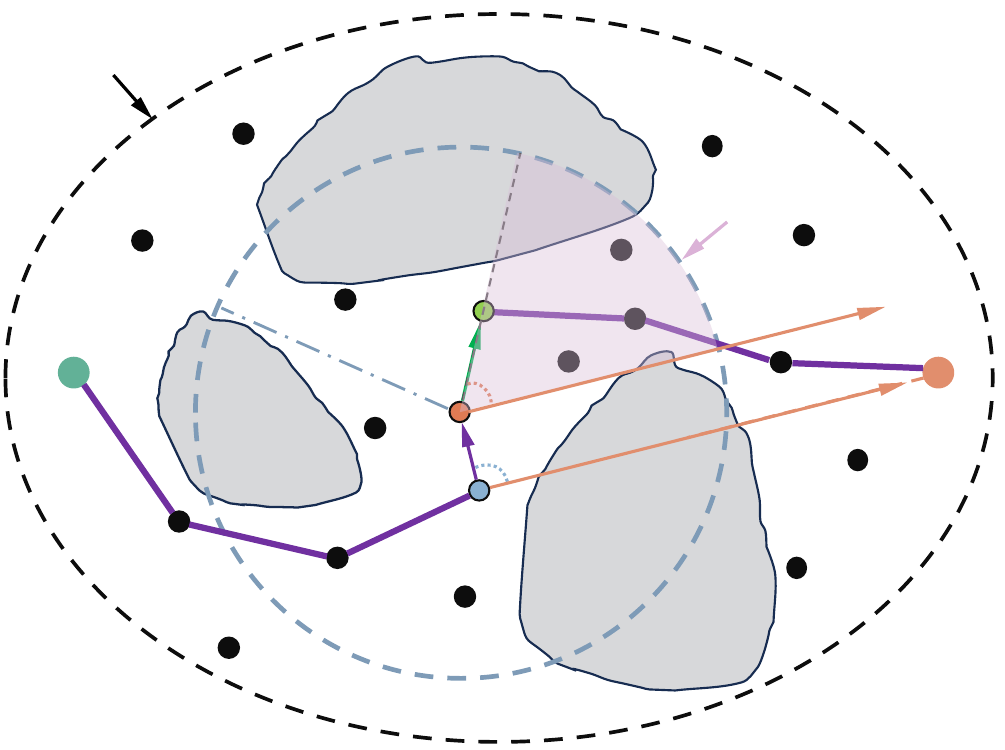}};
    \node at (-3.75,-0.38) {$\mathbf{x}_\text{start}$};
    \node at (-3.35,2.75) {\footnotesize{informed set}};    
    \node at (2.3,1.45) {\color[RGB]{164, 62, 154}\footnotesize{area to explore}};
    \node at (-0.6,2.2) {${X}_\text{obs}$};
    \node at (3.75,-0.38) {$\mathbf{x}_\text{goal}$};
    \node at (3.4,0.83) {\color[RGB]{225, 142, 109}\small$\vec{V}_\textit{goal,shifted}$};
    \node at (2.5,-0.65) {\color[RGB]{225, 142, 109}\small$\vec{V}_\textit{goal}$};
    \node at (-0.65,0.25) {\color[RGB]{0, 176, 80}\small$\vec{V}_\textit{check}$};
    \node at (-0.68,-0.75) {\color[RGB]{112, 48, 160}\small$\vec{V}_\textit{target}$};
    \node at (0.13,-0.5) {\small$\mathbf{x}_\textit{source}$};
    \node at (-0.15,-1.35) {\small$\mathbf{x}_\textit{last}$};
    \node at (0.3,0.75) {\small$\mathbf{x}_\textit{neighbor}$};
    
    \node at (-2.1,0.7) {\color{teal}$r$};
    
    \end{tikzpicture}
    \caption{Last state $\mathbf{x}_\textit{last}$ to source state $\mathbf{x}_\textit{source}$ forms the target vector $\vec{V}_\textit{target}$, $\mathbf{x}_\textit{source}$ to next state $\mathbf{x}_\textit{neighbor}$
    forms vector to check $\vec{V}_\textit{check}$, $\mathbf{x}_\textit{last}$ to goal state forms goal vector $\vec{V}_\textit{goal}$. Additional $\vec{V}_\textit{goal,shifted}$ starts at $\mathbf{x}_\textit{source}$ is shifted from $\mathbf{x}_\textit{last}$ and similar to $\vec{V}_\textit{goal}$. $\vec{V}_\textit{goal,shifted}$ and $\vec{V}_\textit{check}$ created \textit{area to explore} represented in light purple; this will filter out redundant samples to optimize edge evaluation.}
    \label{fig:direction_filter}
    \vspace{-1.7em} 
\end{figure}
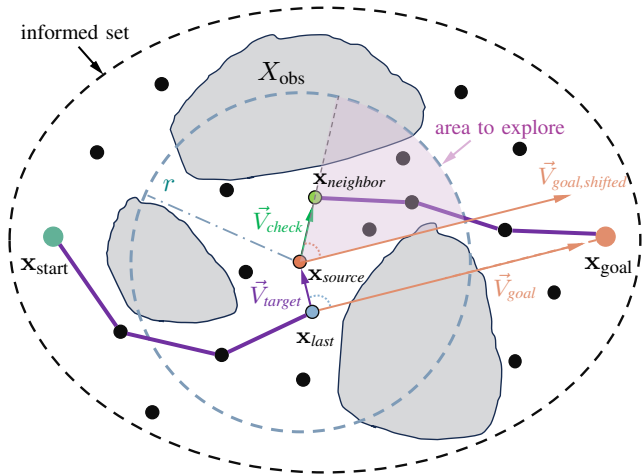

In this paper, we propose Direction Informed Trees (DIT*), a sampling-based path planning algorithm for goal-bias and obstacle avoidance. DIT* integrates a bidirectional guided search strategy to emphasize optimal goal-directed exploration. This is achieved through two approaches: the \textit{direction filter}, which prioritizes goal-oriented directions and filters out less favorable neighbors. The weighted cosine similarity index for \textit{direction cost} heuristic quantifies the directional cost of each direction's proximity to the goal.
We proposed direction filter exploration (Fig.~\ref{fig:direction_filter}), three vectors are established: the \textit{target vector} $\vec{V}_\textit{target}$, the \textit{vector to check} $\vec{V}_\textit{check}$, and the \textit{goal vector} $\vec{V}_\textit{goal}$. The objective is to employ these vectors to impose constraints and establish the direction of the search. The direction is valid (i.e.,~\textit{approaching the goal region}) when $\vec{V}_\textit{check}$ lies between $\vec{V}_\textit{target}$ and $\vec{V}_\textit{goal}$. We measured the deviation from prior state directions by weighted similarity index in high-dimensional vector space.

%
In \textit{$\mathcal{C}$-space}, a transition between a $\mathbf{x}_\textit{source}$ and $\mathbf{x}_\textit{target}$ state is defined by four vectors: $\vec{V}_\textit{target}$, $\vec{V}_\textit{check}$, $\vec{V}_\textit{goal}$, and $\vec{V}_\textit{goal,shifted}$. These correspond to movements from $\mathbf{x}_\textit{last}$ to $\mathbf{x}_\textit{goal}$ and from $\mathbf{x}_\textit{source}$ to $\mathbf{x}_\textit{neighbor}$. When exploring $X_\textit{neighbors}$ set, constraints based on $\vec{V}_\textit{target}$ are applied to filter out unnecessary states and remove nearest neighbor samples that are far from the goal direction, leading to faster initial convergence.

DIT*'s performance is validated through simulated experiments from $\mathbb{R}^4$ to $\mathbb{R}^{16}$ and real-world tasks, including kitchen model robot manipulation~\cite{Bing2022Tnnls} (Fig.~\ref{fig:kitchenModel}) and dual-arm (14-DoF) context-cable routing~\cite{kejia2023cable,Bing2024TPAMI} (Fig.~\ref{fig: dual_arm_setup}). 
Through real-world experiments, DIT* is effective for autonomous robotic manipulation in high-dimensional, cluttered environments.

The contributions of this work are summarized as follows:
\begin{itemize}
    \item \textit{Direction filter constraints:} DIT* introduces a directional filter that integrates generalized vectors to identify area-to-explore. It filters out the less favorable nearest neighbors, thereby improving the convergence rate.
    \item \textit{Direction cost heuristic function:} DIT* incorporates a directional cost heuristic via weighted cosine similarity index to assess edges, directing the goal-biased search.
    \item \textit{Real-world applications and effectiveness:} The behavior of DIT* was shown in real-world single-arm kitchen model and dual-arm cable routing tasks, confirming improvements in path convergence and the solution cost.

\end{itemize}

%% file: sections/sec2_2_statement.tex
\section{Problem Formulation and Notation}


    
We define the problem of optimal planning in a manner associated with the definition provided in~\cite{karaman2011sampling}.
\subsection{Problem Definition}
\textit{Optimal Planning:} Consider a planning problem with the state space $X \subseteq \mathbb{R}^n$. Let $X_{\text{obs}} \subset X$ represent states in collision with obstacles, and $X_{\text{free}} = cl(X \setminus X_{\text{obs}})$ denote the resulting permissible states, where $cl(\cdot)$ represents the \textit{closure} of a set. The initial state is denoted by $\mathbf{x}_{\text{start}} \in X_{\text{free}}$, and the set of desired final states is $X_{\text{goal}} \subset X_{\text{free}}$. A sequence of states $\sigma: [0, 1] \mapsto X$ forms a continuous map (i.e., a collision-free, feasible path), and $\Sigma$ represents the set of all nontrivial paths.
The optimal solution, represented as $\sigma^*$, corresponds to the path that minimizes a selected cost function $c: \Sigma \mapsto \mathbb{R}_{\geq 0}$. This path connects the initial state $\mathbf{x}_{\text{start}}$ to any goal state $\mathbf{x}_{\text{goal}} \in X_{\text{goal}}$ through the free space:
\begin{equation}
\begin{split}
    \sigma^* &= \arg \min_{\sigma \in \Sigma} \left\{ c(\sigma) \middle| \sigma(0) = \mathbf{x}_{\text{start}}, \sigma(1) \in \mathbf{x}_{\text{goal}}, \right. \\
    &\qquad\qquad \left. \forall t \in [0, 1], \sigma(t) \in X_{\text{free}} \right\}.
\end{split}
\end{equation}
where $\mathbb{R}_{\geq 0}$ denotes non-negative real numbers. The cost of the optimal path is $c^*$.
Considering a discrete set of states, $X_{\text{samples}} \subset X$, as a graph where edges are determined algorithmically by a transition function, we can describe its properties using a probabilistic model implicit dense RGGs when these states are randomly sampled, i.e., $X_{\text{samples}} = \{ \mathbf{x} \sim \mathcal{U}(X) \}$, as discussed in~\cite{penrose2003random}.




\subsection{Notation}
The state space of the planning problem is denoted by $X \subseteq \mathbb{R}^n$, where $n \in \mathbb{N}$. The start point is represented by $\mathbf{x}_{\textit{start}} \in X$, and the goals are denoted by $X_{\textit{goal}} \subset X$. The sampled states are denoted by $X_{\textit{sampled}}$. The forward and reverse search trees are represented by $\mathcal{F} = (V_\mathcal{F}, E_\mathcal{F})$ and $\mathcal{R} = (V_\mathcal{R}, E_\mathcal{R})$, respectively. The vertices in these trees, denoted by $V_\mathcal{F}$ and $V_\mathcal{R}$, are associated with valid states. The edges in the forward tree, $E_\mathcal{F} \subseteq V_\mathcal{F} \times V_\mathcal{F}$, represent valid connections between states, while the edges in the reverse tree, $E_\mathcal{R} \subseteq V_\mathcal{R} \times V_\mathcal{R}$, may traverse invalid regions (i.e., sparse collision detection) of the problem domain. An edge comprises a source state, $\mathbf{x}_s$, and a target state, $\mathbf{x}_t$, denoted as $(\mathbf{x}_s, \mathbf{x}_t)$. $\mathcal{Q_F}$ and $\mathcal{Q_R}$ designate the edge queue for the forward search and reverse search, respectively. An admissible estimate for the cost of a path is $\hat{f}: X \rightarrow [0, \infty)$, characterizing the informed set $X_{\hat{f}}$.

\textit{DIT*-specific Notation:}
We define a generalized vector, $\mathcal{V}\textit{ec} \subseteq V_{\mathcal{F}}$, representing an edge vector with directions, where $\mathbf{x}_\mathrm{s}$ and $\mathbf{x}_\mathrm{t} := \mathbf{x}_\textit{neighbor} \in X_\textit{neighbors}$ denote  source and target states within neighbors as $V_{\mathcal{F}}(\mathbf{x}_\mathrm{s})$ and $V_{\mathcal{F}}(\mathbf{x}_\mathrm{t})$. Directional cost heuristics between states, $\bar{dc}: X \times X \rightarrow [0, \infty)$, assess path deviation costs, considering angular divergence from the optimal route. $\bar{dg}: X \rightarrow [0, \infty)$ measures deviation from the goal, defined as $\bar{dg}(\mathbf{x}) := \bar{dg}(\mathbf{x}, \mathbf{x}_{\textit{goal}})$, and $\bar{ds}: X \rightarrow [0, \infty)$ assesses deviation from the start, $\bar{ds}(\mathbf{x}) := \bar{ds}(\mathbf{x}, \mathbf{x}_{\textit{start}})$. These heuristics, though potentially inadmissible, may overestimate directional deviation costs.

%% file: sections/sec3_formulation.tex
\section{Direction Informed Trees (DIT*)}
In this section, we first introduce the concept of direction filter in forward search. Next, we illustrate the direction cost heuristic for best edge. Finally, we prove that DIT* guarantees probabilistic completeness and asymptotic optimality.

%% file: sections/sec4_analysis.tex
\subsection{Direction Filter Constrains}\label{sec:direction_filter}
\begin{algorithm}[t!]
\DontPrintSemicolon
\SetAlgoNlRelativeSize{-1}
\small
\caption{\small\text{DIT* - direction filter}}
\label{alg:combined_filter}
\SetKwInOut{Input}{Input}
\SetKwInOut{Output}{Output}
\SetKwFunction{generalVec}{generalVec}
\SetKwFunction{normalize}{normalize}
\SetKwFunction{wgtCosSim}{wgtCosSim}
\SetKwFunction{isVectorBetween}{isVectorBetween}
\SetKwFunction{calcDirCost}{calcDirCost}
\SetKwFunction{filterNeighbors}{filterNeighbors}

\SetKwProg{Function}{Function}{}{}

\Input{last state $\mathbf{x}_{\textit{last}}$, source state $\mathbf{x}_{\textit{source}}$, \\goal state $\mathbf{x}_{\textit{goal}}$, neighbors set $X_{\textit{neighbors}}$}
\Output{filtered neighbors $X_{\textit{neighbors}}$}
$\mathcal{V}\textit{ec}_{\textit{target}} \gets \generalVec(\mathbf{x}_{\textit{last}}, \mathbf{x}_{\textit{source}})$\;
$\mathcal{V}\textit{ec}_{\textit{goal}} \gets \generalVec(\mathbf{x}_{\textit{last}}, \mathbf{x}_{\textit{goal}})$\;

\ForAll{$\mathbf{x}_{\text{neighbor}} \in X_{\text{neighbors}} \subseteq V_{\mathcal{F}}$}{
    $\mathcal{V}\textit{ec}_{\textit{check}} \gets \generalVec(\mathbf{x}_{\textit{source}}, \mathbf{x}_{\textit{neighbor}})$\;

    \If(\label{line:safe}){$\lvert \mathcal{V}\textit{ec}_{\textit{target}} \rvert < \epsilon$ \textbf{or} $\lvert \mathcal{V}\textit{ec}_{\textit{goal}} \rvert < \epsilon$ \textbf{or} $\lvert \mathcal{V}\textit{ec}_{\textit{check}} \rvert < \epsilon$}{
        \Comment{$\epsilon$ is a judgment of non-zero vector for safety}\\
        \Return False\;
    }

    $\hat{\mathcal{V}\textit{ec}}_{\textit{target}}, \hat{\mathcal{V}\textit{ec}}_{\textit{goal}}, \hat{\mathcal{V}\textit{ec}}_{\textit{check}} \gets \normalize(\mathcal{V}\textit{ec}_{\textit{target}}, \mathcal{V}\textit{ec}_{\textit{goal}}, \mathcal{V}\textit{ec}_{\textit{check}})$\;

    $\Phi_\textit{check}  \gets \wgtCosSim(\hat{\mathcal{V}\textit{ec}}_{\textit{check}}, \hat{\mathcal{V}\textit{ec}}_{\textit{goal}})$ \Comment{index $\Phi$}\\
    $\Phi_\textit{target} \gets \wgtCosSim(\hat{\mathcal{V}\textit{ec}}_{\textit{target}}, \hat{\mathcal{V}\textit{ec}}_{\textit{goal}})$\;

    \eIf{\textcolor{purple}{$\Phi_\text{check} \geq \Phi_\text{target}$}}{
        \textcolor{purple}{$X_{\text{neighbors}} \gets X_{\text{neighbors}} \setminus \mathbf{x}_{\textit{neighbor}}$} \Comment{filter neighbors}\\

    }{
        \textcolor{purple}{$\bar{dc}(\mathbf{x}_{\mathrm{s}}, \mathbf{x}_{\mathrm{t}}) \leftarrow$ \calcDirCost{$\Phi_\textit{check}, \Phi_\textit{target},  \mathbf{x}_{\textit{neighbor}}$}}
    }
}

\Return $X_{\textit{neighbors}}$
\end{algorithm}
In path planning, collision-checking is the most computationally expensive operation. To optimize efficiency, we introduce a filtering mechanism (Alg.~\ref{alg:combined_filter}) that selectively excludes less promising neighboring states, thereby reducing the number of edges that need to be established and checked. Specifically, we define edges as \textit{generalized vectors} $\mathcal{V}\textit{ec} \coloneqq \mathcal{V}\textit{ec}(\mathbf{x}_s, \mathbf{x}_t)$, where $\mathcal{V}\textit{ec} \in \mathbb{R}^{n \times 1}$, with all states being $n$-dimensional. This approach offers the advantage of representing each state in the forward search graph as a vector derived from its numerical value:
%
\begin{equation}
\mathcal{V}\textit{ec} := V_{\mathcal{F}}\left( \mathbf{x}_s \right) \,\,-\,\,V_{\mathcal{F}}\left( \mathbf{x}_t \right),
\label{eq:label}
\end{equation}

The mathematical property and orientation of $\mathcal{V}\textit{ec}$ are defined as $\left\| \mathcal{V}\textit{ec} \right\| := \sqrt{\sum_{i=1}^n \left( \mathcal{V}\textit{ec}_i \right)^2}$, $\hat{\mathcal{V}\textit{ec}} := \mathcal{V}\textit{ec}/{\left\| \mathcal{V}\textit{ec} \right\|}$, where $\hat{\mathcal{V}\textit{ec}} \in \mathbb{R}^{n \times 1}$. Moreover, the measure of the weighted cosine similarity, $\texttt{wgtCosSim}(\cdot)$ is defined as follows:
\begin{equation}
    \Phi_{i} := \frac{\sum_{i=1}^{n}  \left(\hat{\mathcal{V}\textit{ec}}_{1,i} \cdot \hat{\mathcal{V}\textit{ec}}_{2,i}\right)}{\sqrt{\sum_{i=1}^{n} \left(\omega_1 \cdot \hat{\mathcal{V}\textit{ec}}_{1,i}\right)^2} \cdot \sqrt{\sum_{i=1}^{n} \left(\omega_2 \cdot \hat{\mathcal{V}\textit{ec}}_{2,i}\right)^2}},
\end{equation}
%
%
%
where similarity index $\Phi_{i} \in [-1, 1]$ is the direction angle projection, and $\omega_1, \omega_2 \in (0, 1)$ are the prior weight parameters. The normalized generalized vector $\hat{\mathcal{V}\textit{ec}}$ is utilized to quantify the direction projection of edges in the filtering process and to calculate the directional cost attribute of each state (Alg.~\ref{alg:DirectionCost}, line 11). This approach prioritizes exploration toward the direction of the goal and excludes unnecessary distant states during the search, thereby improving efficiency.

DIT* incorporates both admissible and inadmissible heuristics in its exploration. The admissible cost heuristic, $\widehat{h}[\cdot]$, provides a minimum path cost to the goal, while the inadmissible $\bar{h}[\cdot]$ integrates problem-specific knowledge, often overestimating path costs for greater precision.
The inadmissible effort heuristic, $\bar{e}[\cdot]$, estimates the computational effort required for path validation. The admissible heuristic, $\widehat{dc}[\cdot]$, provides a lower bound for optimal directional path costs. In contrast, the inadmissible heuristic $\bar{dc}[\cdot]$ estimates the optimal direction costs. The edge cosine similarity heuristic incorporates path length and direction angle index $\Phi_\textit{i}$.

Therefore, we have associated the cost of the inadmissible direction heuristic with the angle directional projection in vector space, inspired by the concept of sigmoid function:
\begin{equation}
\widehat{dc}_\textit{cost} :=  \left\lfloor \sqrt{  \xi_1 \left( \frac{\left\| \mathcal{V}\textit{ec}_1 \right\|}{1+e^{-\Phi_1}} \right)^2 + \xi_2  \left( \frac{\left\| \mathcal{V}\textit{ec}_2 \right\|}{1+e^{-\Phi_2}} \right)^2 }\right\rfloor,
\label{eq:inadmissible_direction_cost}
\end{equation}
where $\lfloor \cdot \rfloor$ denotes the floor function for integers, $e$ is Euler’s number $e = \sum_{i=0}^\infty \frac{1}{n!}$. The tuning parameter $\xi \in (0,1)$ is to balance exploration and the prior direction projection of $\mathcal{V}\textit{ec}$. The inadmissible direction cost is denoted as $\widehat{dc}_\textit{cost} \in \mathbb{N}$, and it quantifies the directional projection distance of edges.
\begin{algorithm}[t!]
\DontPrintSemicolon
\small
\caption{\small\text{DIT* - direction cost}}
\label{alg:DirectionCost}

\SetAlgoNlRelativeSize{-2}
\SetKwInOut{Input}{Input}
\SetKwInOut{Output}{Output}
\SetKwFunction{generalVec}{generalVec}
\SetKwFunction{getNorm}{getNorm}
\SetKwFunction{wgtCosSim}{wgtCosSim}
\SetKwFunction{setDirCost}{setDirCost}
\SetKwFunction{calcDirCost}{calcDirCost}

\SetAlgoNlRelativeSize{-1}
\SetKwProg{Function}{Function}{}{}
\Input{check angle index $\Phi_\textit{check}$, target angle index $\Phi_\textit{target}$, neighbors states $\mathbf{x}_{\textit{neighbor}}$}
\Output{neighbor edge direction cost $\bar{dc}(\mathbf{x}_{\mathrm{s}}, \mathbf{x}_{\mathrm{t}})$}
\Function{\calcDirCost{$\Phi_\textit{check}, \Phi_\textit{target},  \mathbf{x}_{\textit{neighbor}}$}}{

    \If(\label{alg: threshold}){$|\Phi_\textit{check}|  \approx 1$ \textbf{or} $|\Phi_\textit{target}|  \approx 1$}{
        $\widehat{d}c_\textit{cost} \gets 0$\Comment{ best edge with maximal similarity }\\
        $\bar{dc}(\mathbf{x}_{\mathrm{s}}, \mathbf{x}_{\textit{neighbor}}) \gets \setDirCost(\mathbf{x}_\textit{neighbor}, \widehat{d}c_\textit{cost})$\;
        \KwRet $\bar{dc}(\mathbf{x}_{\mathrm{s}}, \mathbf{x}_{\textit{neighbor}})$\;
    }

    \emph{\label{alg: directionflg1}\color{purple}$\Theta_\textit{target} \gets ||\mathcal{V}\textit{ec}_\textit{target}|| / 
           (1+e^{-\Phi_\textit{target}})$}\\\Comment{$\Theta$ is the scaled direction projection for edges}\\
    \emph{\color{purple}$\Theta_\textit{check} \gets ||\mathcal{V}\textit{ec}_\textit{check}|| / 
           (1+e^{-\Phi_\textit{check}})$\;}

    $\xi_1 \gets 0.6$
    \Comment{tuning parameter for prior measure}\\
    $\xi_2 \gets 0.8$
    \Comment{tuning parameter for explore}\\
    
    $\widehat{d}c_\textit{cost} \gets \lfloor \sqrt{ \xi_1 \cdot (\Theta_\textit{target})^2 +
           \xi_2 \cdot (\Theta_\textit{check})^2 }\rfloor$\\\Comment{scale interval for direction comparison, Eq.~\ref{eq:inadmissible_direction_cost}.}

    $\bar{dc}(\mathbf{x}_{\mathrm{s}}, \mathbf{x}_{\textit{neighbor}}) \gets \setDirCost(\mathbf{x}_\textit{neighbor}, \widehat{d}c_\textit{cost})$\\
    \Return $\bar{dc}(\mathbf{x}_{\mathrm{s}}, \mathbf{x}_{\textit{neighbor}})$
}
\end{algorithm}

We define the search in DIT* as an edge-queue version with cost $\widehat{c}\left( \mathbf{x}_{\mathrm{s}}, \mathbf{x}_{\mathrm{t}} \right)$, effort $\bar{e}\left( \mathbf{x}_{\mathrm{s}}, \mathbf{x}_{\mathrm{t}} \right)$ and directional cost $\bar{dc}\left( \mathbf{x}_{\mathrm{s}}, \mathbf{x}_{\mathrm{t}} \right)$ heuristics. The queue for the reverse search, denoted by $\mathcal{Q}_{\mathcal{R}}$, is ordered lexicographically according to:
\begin{equation}
\mathrm{key}_{\mathcal{R}}^{\mathrm{DIT}^*}\left( \mathbf{x}_{\mathrm{s}},\mathbf{x}_{\mathrm{t}} \right) :=
\begin{cases}
\widehat{h}[\mathbf{x}_{\mathrm{s}}] + \widehat{c}\left( \mathbf{x}_{\mathrm{s}}, \mathbf{x}_{\mathrm{t}} \right) + \widehat{g}\left( \mathbf{x}_{\mathrm{t}} \right),\\
\bar{e}[\mathbf{x}_{\mathrm{s}}] + \bar{e}\left( \mathbf{x}_{\mathrm{s}}, \mathbf{x}_{\mathrm{t}} \right) + \bar{d}\left( \mathbf{x}_{\mathrm{t}} \right),\\
\widehat{ds} [\mathbf{x}_{\mathrm{s}}] + \bar{dc}\left( \mathbf{x}_{\mathrm{s}}, \mathbf{x}_{\mathrm{t}} \right) + \widehat{dg}\left( \mathbf{x}_{\mathrm{t}} \right).\\
\end{cases}
\end{equation}
where the admissible cost $\widehat{h}[\cdot]$ and direction heuristics $\widehat{ds}[\cdot]$ are computed using a priori admissible cost $\widehat{c}[\cdot, \cdot]$ and direction cost heuristics $\widehat{dc}[\cdot, \cdot]$, ensuring heuristic admissibility. Conversely, the inadmissible cost $\bar{h}[\cdot]$, effort $\bar{e}[\cdot]$, and direction cost heuristics $\bar{ds}[\cdot]$ use inadmissible a priori values $\bar{c}(\cdot, \cdot)$, $\bar{e}(\cdot,\cdot)$, and $\bar{dc}(\cdot, \cdot)$. 
These components are integrated into the key function, $\mathrm{key}_{\mathcal{R}}^{\mathrm{DIT}^*}$, which represents the total solution cost, computational effort, and direction cost associated with navigating a path through an edge. The first part of the key ensures the admissibility of the calculated cost heuristic. The second part of the key ensures tiebreaks in favor of lower estimated effort, which is essential if only the trivially admissible cost heuristic is available, i.e., $\forall \mathbf{x}, \mathbf{x}^{\prime} \in X, \widehat{c}\left(\mathbf{x}, \mathbf{x}^{\prime}\right) \equiv \widehat{g}(\mathbf{x}) \equiv 0$. The \textit{direction cost} serves as the third part of the key value hierarchy when effort heuristic values are comparable (i.e., $\forall \mathbf{x}, \mathbf{x}^{\prime}, \mathbf{x}^{\prime\prime} \in X, | \widehat{e}\left(\mathbf{x}, \mathbf{x}^{\prime}\right) - \widehat{e}\left(\mathbf{x}^{\prime},\mathbf{x}^{\prime\prime} \right) | \leq \mu \in \mathbb{N}$). We prioritize the third key in cases of under-threshold computational efforts.

Overall, we integrate admissible heuristics to ensure a conservative, shortest-path search, while employing inadmissible heuristics to expedite goal discovery by allowing for potential overestimations. We prioritize edges with lower directional costs as the more optimal choice in the path planning process.

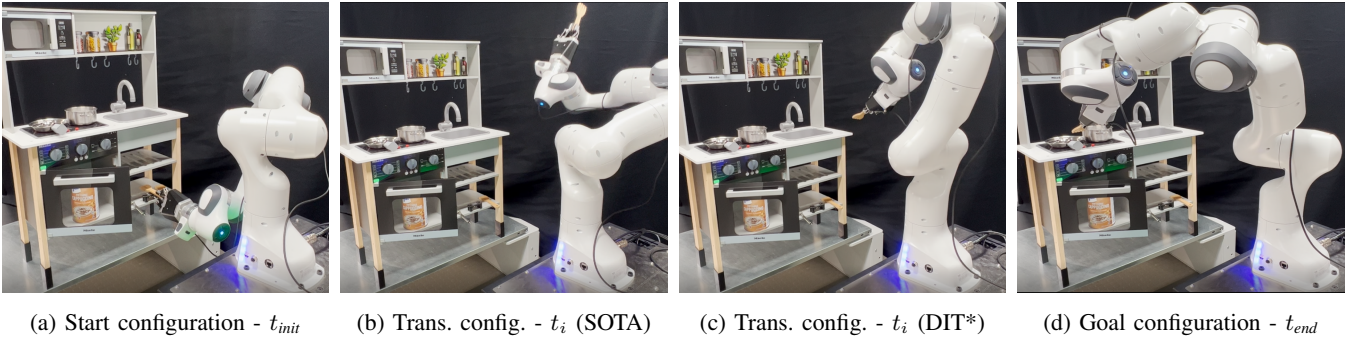
\begin{figure*}[t!]
    \centering
    \begin{tikzpicture}
    \footnotesize
    \node[inner sep=0pt] (russell) at (-8.0,0)
    {\includegraphics[width=0.244\textwidth]{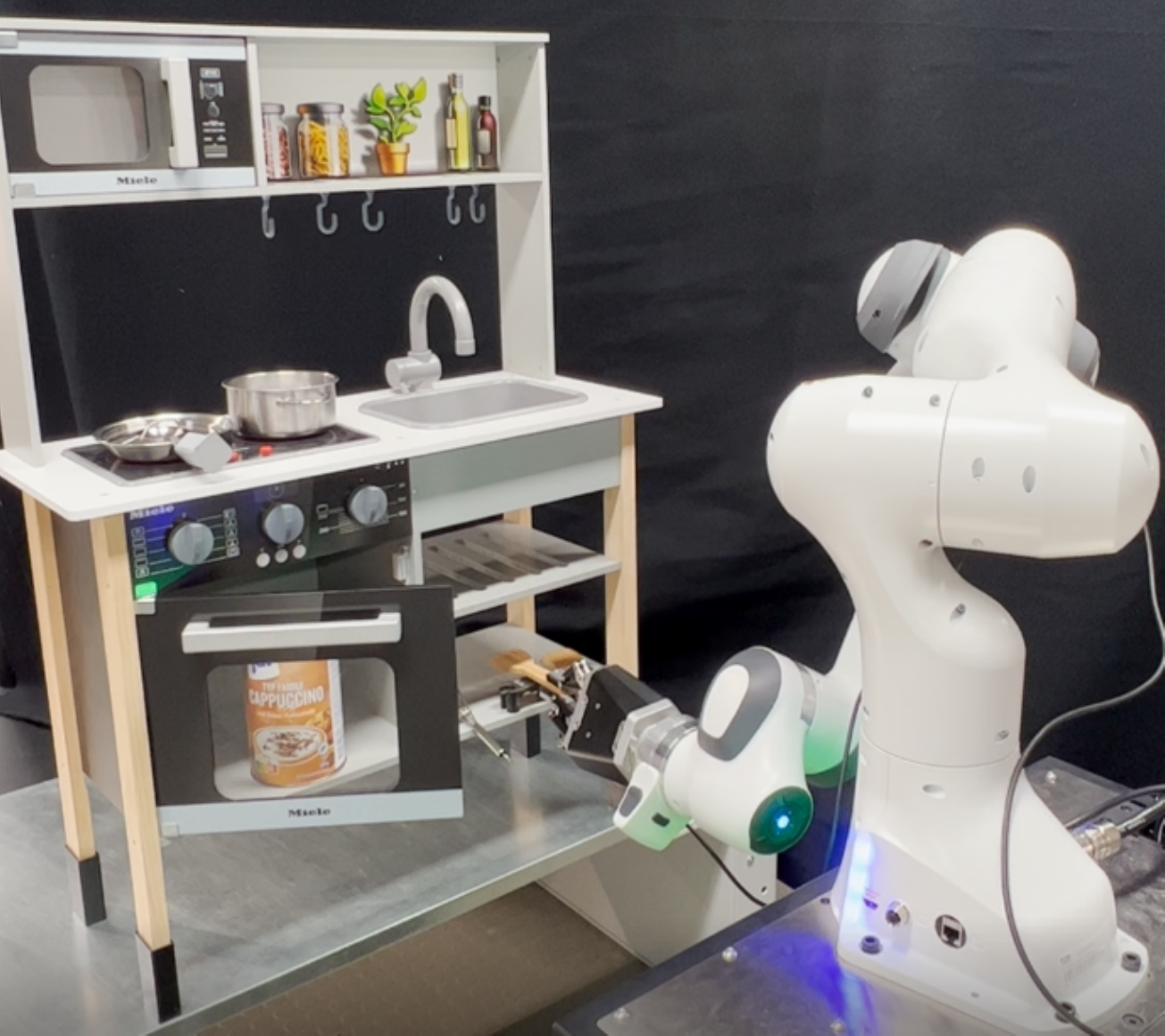}};
    \node[inner sep=0pt] (russell) at (-3.5,0)
    {\includegraphics[width=0.244\textwidth]{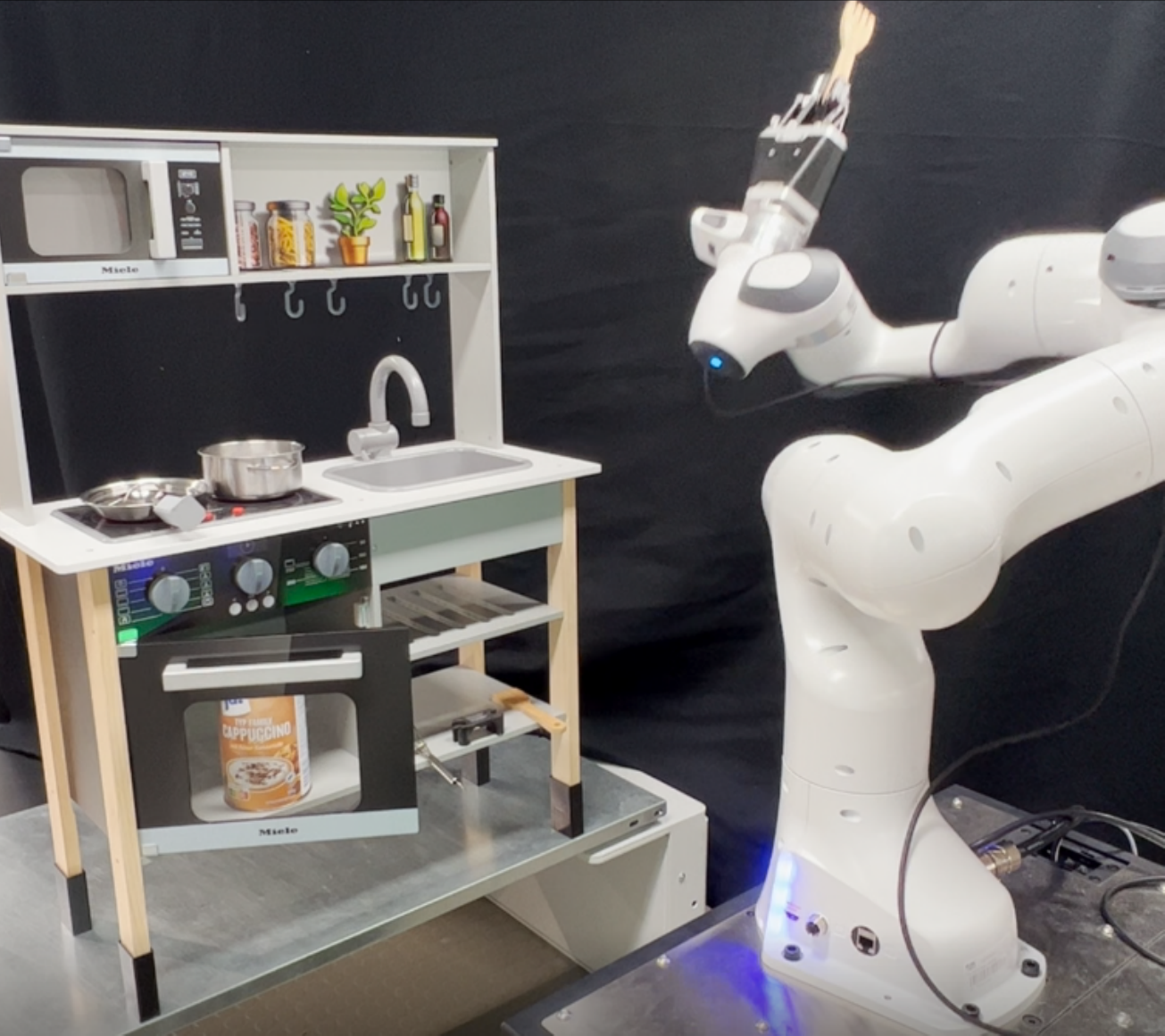}};
    \node[inner sep=0pt] (russell) at (1.0,0)
    {\includegraphics[width=0.244\textwidth]{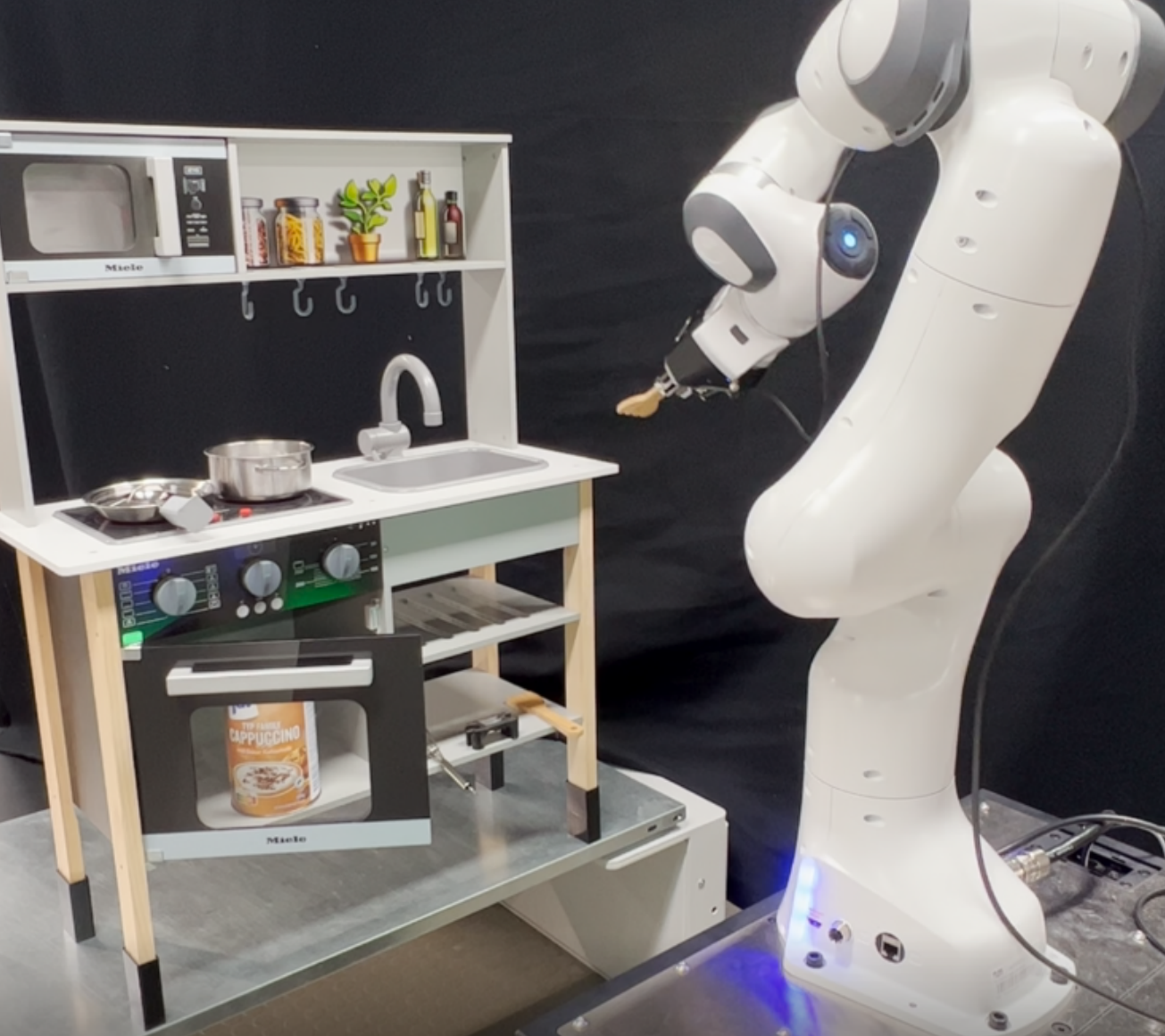}};
    \node[inner sep=0pt] (russell) at (5.5,0)
    {\includegraphics[width=0.244\textwidth]{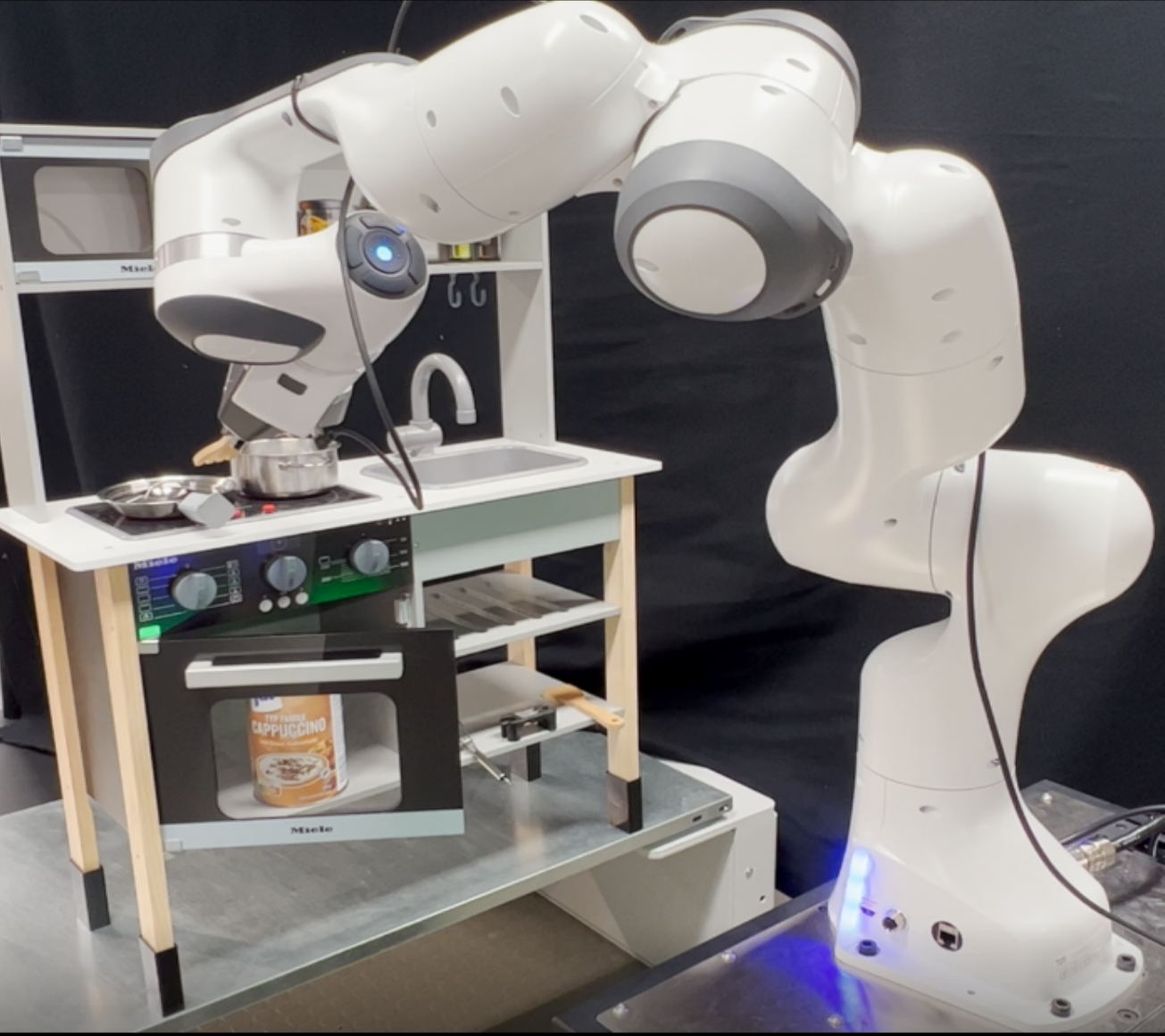}};
    
    \node at (-8.0,-2.35) {\small (a) Start configuration - $t_\textit{init}$};
    \node at (-3.5,-2.35) {\small (b) Trans. config.  - $t_i$ (SOTA)};
    \node at (1.0,-2.35) {\small (c) Trans. config.  - $t_i$ (DIT*)};
    \node at (5.5,-2.35) {\small (d) Goal configuration - $t_\textit{end}$};

    \end{tikzpicture}
    \caption{Illustrates the real-world scenarios of a manipulator (7-DoF) for the kitchen cooking task, (a) shows the start configuration of the manipulator in position to grasp a yellow fork from the lower drawer, (b) and (c) shows the transition configuration of SOTA and DIT* planner at the same time stamp, (d) shows the goal configuration of the arm in position to hold the yellow fork near the pan.}
    \label{fig:kitchenModel}
    \vspace{-1.5em}
\end{figure*}

\subsection{Direction Cost Heuristic}
\begin{algorithm}[t]

\SetAlgoNlRelativeSize{-1}
\caption{\small\text{DIT* - select best forward edge from $\mathcal{Q}_{\mathcal{F}}$}}\label{alg:get_best_forward_edge}

\SetAlgoNlRelativeSize{-2}
\SetKwInOut{Input}{Input}
\SetKwInOut{Output}{Output}
\SetKwFunction{length}{length}
\DontPrintSemicolon
\small
\SetAlgoNlRelativeSize{-1}
\SetKwProg{Function}{Function}{}{}
\SetKwInOut{Require}{Require}

\Input{forward queue $\mathcal{Q}_{\mathcal{F}}$, direction cost $\bar{dc}(\mathbf{x}_{\mathrm{s}}, \mathbf{x}_{\mathrm{t}})$}
\Output{best forward edge $E_\mathcal{F}(\mathbf{x}_{\mathrm{s}}, \mathbf{x}_{\mathrm{t}})$}





$L \gets \emptyset$ \Comment{initialize list $L$ to store edges}\\
$\bar{r}_{\min} \gets \infty$ 
\Comment{initialize the minimum effort}\\
$\bar{sd}_{\min} \gets \infty$ 
\Comment{initialize the minimum direction cost}\\
\emph{\color{purple}$(\mathbf{x}_{\mathrm{s}}^{\bar{sd}}, \mathbf{x}_{\mathrm{t}}^{\bar{sd}}) \leftarrow \underset{(\mathbf{x}_{\mathrm{s}}, \mathbf{x}_{\mathrm{t}})\in \mathcal{Q}_{\mathcal{F}}}{\operatorname{argmin}} \{ ds_{\mathcal{F}}(\mathbf{x}_{\mathrm{s}}) + \bar{dc}(\mathbf{x}_{\mathrm{s}}, \mathbf{x}_{\mathrm{t}}) + \bar{dg}[\mathbf{x}_{\mathrm{t}}] \}$\;}
\ForAll(\Comment{loop over all edges in queue}){$E_\mathcal{F}(\mathbf{x}_{\mathrm{s}}, \mathbf{x}_{\mathrm{t}}) \in \mathcal{Q}_{\mathcal{F}}$}{
    \uIf{$\bar{e}(\mathbf{x}_{\mathrm{s}}, \mathbf{x}_{\mathrm{t}}) < \bar{r}_{\min}$}{
        $L \gets E_\mathcal{F}(\mathbf{x}_{\mathrm{s}}, \mathbf{x}_{\mathrm{t}})$ \Comment{add edge to list $L$}\\
        $\bar{r}_{\min} \gets \bar{e}(\mathbf{x}_{\mathrm{s}}, \mathbf{x}_{\mathrm{t}})$\;
    }
    \ElseIf(\Comment{edge has more effort}){$\bar{e}(\mathbf{x}_{\mathrm{s}}, \mathbf{x}_{\mathrm{t}}) \geq \bar{r}_{\min}$}{

        $L \gets L \setminus E_\mathcal{F}(\mathbf{x}_{\mathrm{s}}, \mathbf{x}_{\mathrm{t}})$ \Comment{clear edge from list $L$}
    }
}

\uIf(\Comment{if more edges in the list $L$}){$\length(L) > 1$}{
    \ForAll{$E_\mathcal{F}(\mathbf{x}_{\mathrm{s}}, \mathbf{x}_{\mathrm{t}}) \in L$}{
        \emph{{\If{\color{purple}$\bar{dc}(\mathbf{x}_{\mathrm{s}}, \mathbf{x}_{\mathrm{t}}) < \bar{sd}_{\min}$}{
        \emph{\color{purple}{$\bar{sd}_{\min} \gets \bar{dc}(\mathbf{x}_{\mathrm{s}}, \mathbf{x}_{\mathrm{t}})$}\color{black}{\Comment{Eq.~\ref{eq:minimalDirCos}}}}\\
        \emph{\color{purple}$(\mathbf{x}_{\mathrm{s}}^{\bar{sd}}, \mathbf{x}_{\mathrm{t}}^{\bar{sd}}) \gets E_\mathcal{F}(\mathbf{x}_{\mathrm{s}}, \mathbf{x}_{\mathrm{t}})$\color{black}{\Comment{Eq.~\ref{eq:edgeMinDir}}}}\\
        }}}
    }
    \Return $(\mathbf{x}_{\mathrm{s}}^{\bar{sd}}, \mathbf{x}_{\mathrm{t}}^{\bar{sd}})$\Comment{best goal-towards direction edge}
}
\Else(\Comment{if only one edge in the list $L$}){
    \Return the only edge in $L$\;
}

\end{algorithm}
From Section~\ref{sec:direction_filter}, DIT* utilizes a queue-based forward search strategy with cost, effort, and direction cost heuristics that are calculated or estimated from reverse search. The direction cost heuristic helps determine the search direction efficiently. Upon finding an initial solution, DIT* employs admissible and inadmissible direction cost heuristics.

In its forward search, DIT* prioritizes its queue based on four factors: (i) projected lower bound of the optimal solution cost, (ii) estimate of this optimal cost, (iii) minimal remaining effort to validate a solution within suboptimality bounds, and (iv) minimal solution directional cost within these bounds. We focus on the optimal direction cost in this section.

\textit{1) Optimal direction cost bound:} A lower bound on the optimal solution direction cost in the current RGG approximation is computed as:
\begin{equation} 
\min _{\left(\mathbf{x}_{\mathrm{s}}, \mathbf{x}_{\mathrm{t}}\right) \in \mathcal{Q}_{\mathcal{F}}} \widehat{sd}\left(\mathbf{x}_{\mathrm{s}}, \mathbf{x}_{\mathrm{t}}\right),
\label{eq:optimal direction cost}
\end{equation}
where $\widehat{sd}: X_{\text {sampled }} \times X_{\text{sampled }}\rightarrow[0, \infty)$ is the estimation direction cost through an edge, denoted as:
\begin{equation} 
\widehat{sd}\left(\mathbf{x}_{\mathrm{s}}, \mathbf{x}_{\mathrm{t}}\right):=ds_{\mathcal{F}}\left(\mathbf{x}_{\mathrm{s}}\right)+\widehat{dc}\left(\mathbf{x}_{\mathrm{s}}, \mathbf{x}_{\mathrm{t}}\right)+\widehat{dg}\left[\mathbf{x}_{\mathrm{t}}\right],
\end{equation}
where $ds_{\mathcal{F}}\left(\mathbf{x}_{\mathrm{s}}\right)$ is the direction cost to come to the source of the edge, $\widehat{dc}\left(\mathbf{x}_{\mathrm{s}}, \mathbf{x}_{\mathrm{t}}\right)$ is the admissible direction cost heuristic of the edge, and $\widehat{dg}\left[\mathbf{x}_{\mathrm{t}}\right]$ is the calculated admissible direction cost heuristic to go from the target of the edge. At least one edge in the forward queue has an optimally connected source state. The edge with the smallest $\widehat{s}$-value in the queue is a lower bound on the optimal solution direction cost in the current RGG approximation. It is denoted as:

\begin{equation}  
    \left(\widehat{\mathbf{x}_{\mathrm{s}}^{sd}}, \widehat{\mathbf{x}_{\mathrm{t}}^{sd}}\right):=\underset{\left(\mathbf{x}_{\mathrm{s}}, \mathbf{x}_{\mathrm{t}}\right) \in \mathcal{Q}_{\mathcal{F}}}{\operatorname{argmin}} \widehat{sd}\left(\mathbf{x}_{\mathrm{s}}, \mathbf{x}_{\mathrm{t}}\right).
\end{equation}

\textit{2) Optimal direction cost estimate:} A more accurate, but potentially inadmissible, estimate of the optimal direction cost in the current RGG approximation is computed as:
\begin{equation}
\label{eq:minimalDirCos}
\min _{\left(\mathbf{x}_{\mathrm{s}}, \mathbf{x}_{\mathrm{t}}\right) \in \mathcal{Q}_{\mathcal{F}}} \bar{sd}\left(\mathbf{x}_{\mathrm{s}}, \mathbf{x}_{\mathrm{t}}\right),
\end{equation}
where $\bar{sd}: X_{\text {sampled }} \times X_{\text {sampled }} \rightarrow[0, \infty)$ is also an estimate of the direction cost through an edge but with the inadmissible heuristics $\bar{dc}$ and $\bar{dg}$ instead of the admissible heuristics $\widehat{dc}$ and $\widehat{dg}$. It is defined as:

\begin{equation}
\bar{sd}\left(\mathbf{x}_{\mathrm{s}}, \mathbf{x}_{\mathrm{t}}\right):=ds_{\mathcal{F}}\left(\mathbf{x}_{\mathrm{s}}\right)+\bar{dc}\left(\mathbf{x}_{\mathrm{s}}, \mathbf{x}_{\mathrm{t}}\right)+\bar{dg}\left[\mathbf{x}_{\mathrm{t}}\right],
\end{equation}

This estimate is often more accurate than the admissible lower bound because it can use information that may overestimate the true cost. The edge that leads to this possibly inadmissible estimate of optimal direction cost is denoted as:
\begin{equation}
\label{eq:edgeMinDir}
\left(\mathbf{x}_{\mathrm{s}}^{\bar{sd}}, \mathbf{x}_{\mathrm{t}}^{\bar{sd}}\right):=\underset{\left(\mathbf{x}_{\mathrm{s}}, \mathbf{x}_{\mathrm{t}}\right) \in \mathcal{Q}_{\mathcal{F}}}{\operatorname{argmin}} \bar{sd}\left(\mathbf{x}_{\mathrm{s}}, \mathbf{x}_{\mathrm{t}}\right).
\end{equation}

DIT* begins by examining the queue to identify potential edges that could improve the current solution. The selection criteria are detailed in Alg.~\ref{alg:get_best_forward_edge}. 
%
When efforts are comparable, DIT* selects the \textit{edge with minimal directional costs} $(\mathbf{x}_{\mathrm{s}}^{\bar{sd}}, \mathbf{x}_{\mathrm{t}}^{\bar{sd}})$, which has the lowest estimated directional cost. This prioritized edge is expected to lead toward a solution within the established suboptimality bounds.
%
%

%
\subsection{Probabilistic Completeness and Asymptotic Optimality}\label{subsec: approx.}
Most informed tree-based path planning algorithms are probabilistically complete and asymptotically optimal, and DIT* guarantees these properties. DIT* utilizes uniform sampling, ensuring that as the number of iterations $i$ approaches infinity, the entire state space is explored:
\begin{equation}
\lim_{i \to \infty} \mathbb{P} (\{V_\mathcal{F}\cup V_\mathcal{R}\} \cap X_{\text{goal}} \neq \emptyset) = 1,
\end{equation}
this indicating that if a feasible path exists, DIT* will find it, thus guaranteeing probabilistic completeness.
DIT* also employs the same \textit{choose parent} and \textit{rewire} strategies as EIT*. The rewiring radius $r(q)$ in these processes satisfy:
\begin{equation}
\label{eqn:radius r}
    r(q) > \eta \left(2 \left(1 + \frac{1}{n}\right)\left(\frac{\lambda(X_{\hat{f}})}{\zeta_n}\right) \left( \frac{\log(q)}{q}\right)\right)^{\frac{1}{n}},
\end{equation}
where $q$ is the number of sampled states, $\eta > 1$ is a tuning parameter, $n$ is the dimensionality of the problem domain, $\lambda(X_{\hat{f}})$ is the Lebesgue measure of informed set $X_{\hat{f}}$ and $\zeta_n$ is the volume of a unit ball in \textit{$\mathcal{C}$-space}.

By Lemmas 56, 71, and 72 in~\cite{karaman2011sampling}, the following holds:
\begin{equation}
\mathbb{P} \left(\limsup_{q \to \infty} \min_{\sigma \in \Sigma_q} \left\{ c(\sigma) \right\} = c^*\right) = 1,
\end{equation}
where $q$ is the number of samples, $\Sigma_q \subset \Sigma$ is the set of valid paths found by the planner, $c: \Sigma \rightarrow [0, \infty)$ is the cost function, and $c^*$ is the optimal cost. This guarantees that DIT* can find an optimal path as the number of iterations increases, ensuring asymptotic optimality.

%% file: sections/sec5_experiment.tex
\section{Experimental Results}\label{sec:experi}
\begin{figure}[t!]
    \centering
    \begin{tikzpicture}
    \node[inner sep=0pt] (russell) at (-4.0,0.0)
    {\includegraphics[width=0.24\textwidth]{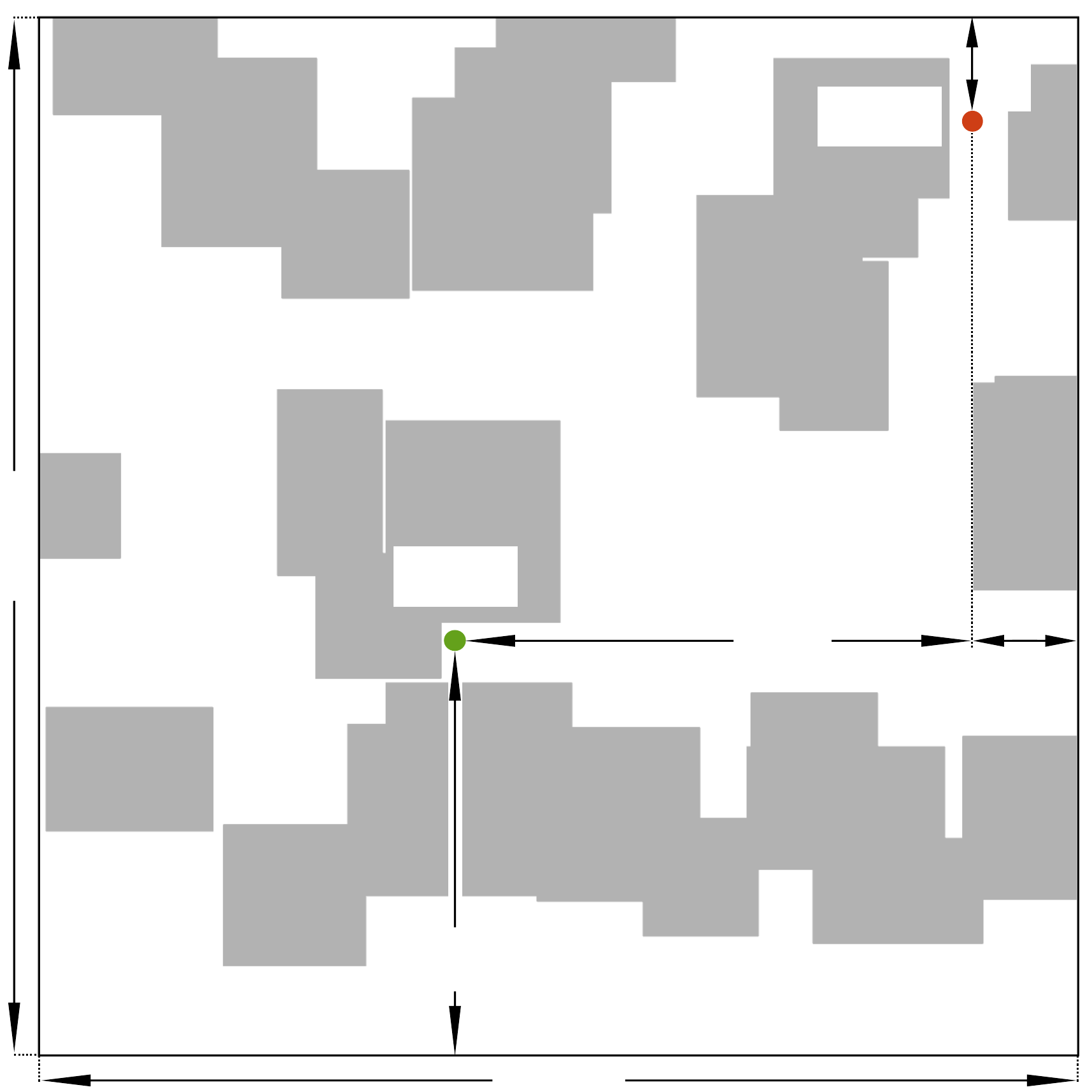}};
    \node[inner sep=0pt] (russell) at (0.25,0.0)
    {\includegraphics[width=0.24\textwidth]{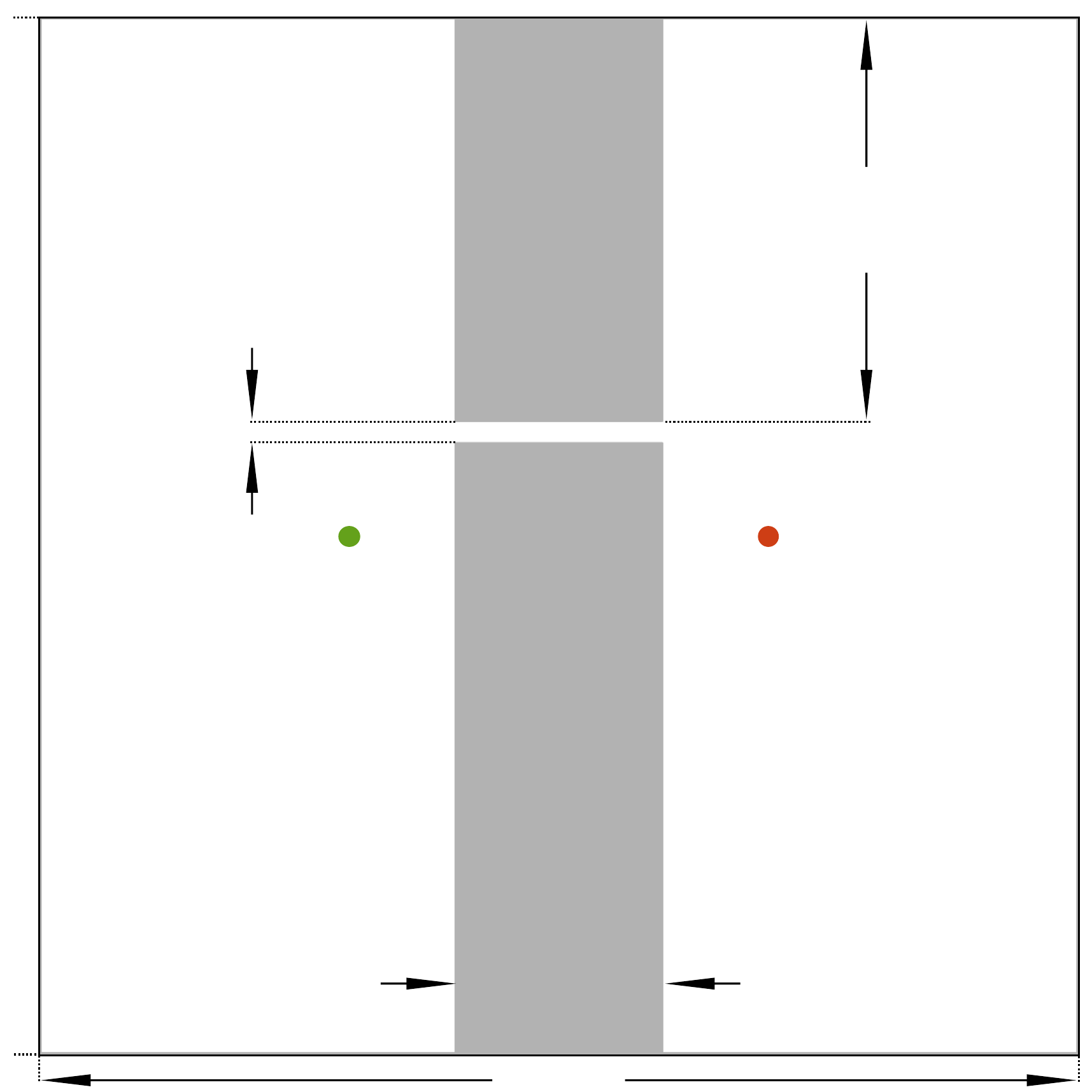}};
    \scriptsize
    
    \node at (-3.08,-0.38) {0.5};
    \node at (-2.12,-0.52) {0.1};
    \node [rotate=90] at (-2.21,1.9) {0.1};
    \node [rotate=90] at (-4.37,-1.61) {0.4};
    
    \node [rotate=90] at (-6.12,0.05) {1.0};
    \node at (-4.0,-2.13) {1.0};
    \node at (-4.36,-0.12) {\color{teal} Start};
    \node at (-2.7,1.68) {\color{purple} Goal};
    
    \node at (1.28,-0.5) {(0.7, 0.5)};
    \node at (-0.68,-0.5) {(0.3, 0.5)};
    \node at (-0.56,-1.7) {0.2};
    \node [rotate=90] at (-1.05,0.49) {0.02};
    \node [rotate=90] at (1.498,1.282) {0.38};
    \node at (0.25,-2.13) {1.0};
    \node at (-0.68,-0.2) {\color{teal} Start};
    \node at (1.28,-0.2) {\color{purple} Goal};

    \node at (-4.0,-2.51) {\small (a) Random Rectangles (RR)};
    \node at (0.25,-2.51) {\small (b) Narrow Passage (NP)};
    \end{tikzpicture}
    \caption{Two snapshots of 2D depiction to demonstrate the simulated planning challenges (Section~\ref{sec:experi}). The state space indicated as $X \subset \mathbb{R}^n$, is confined to a hypercube with a width of one in both instances of the problem. To be precise, we carried out ten unique implementations of the random rectangles and narrow passage experiment, with the results presented in the Fig.~\ref{fig: result}.}
    \label{fig: testEnv}
    \vspace{-1.7em} 
\end{figure}
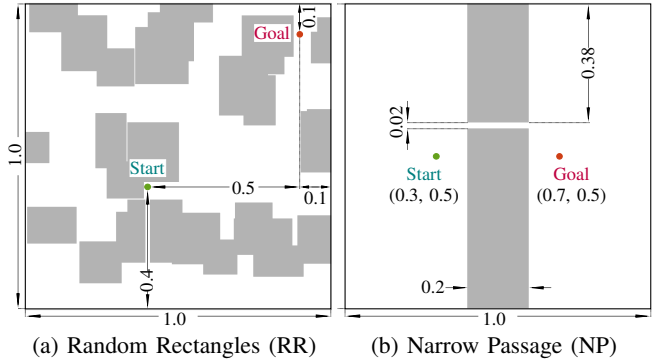
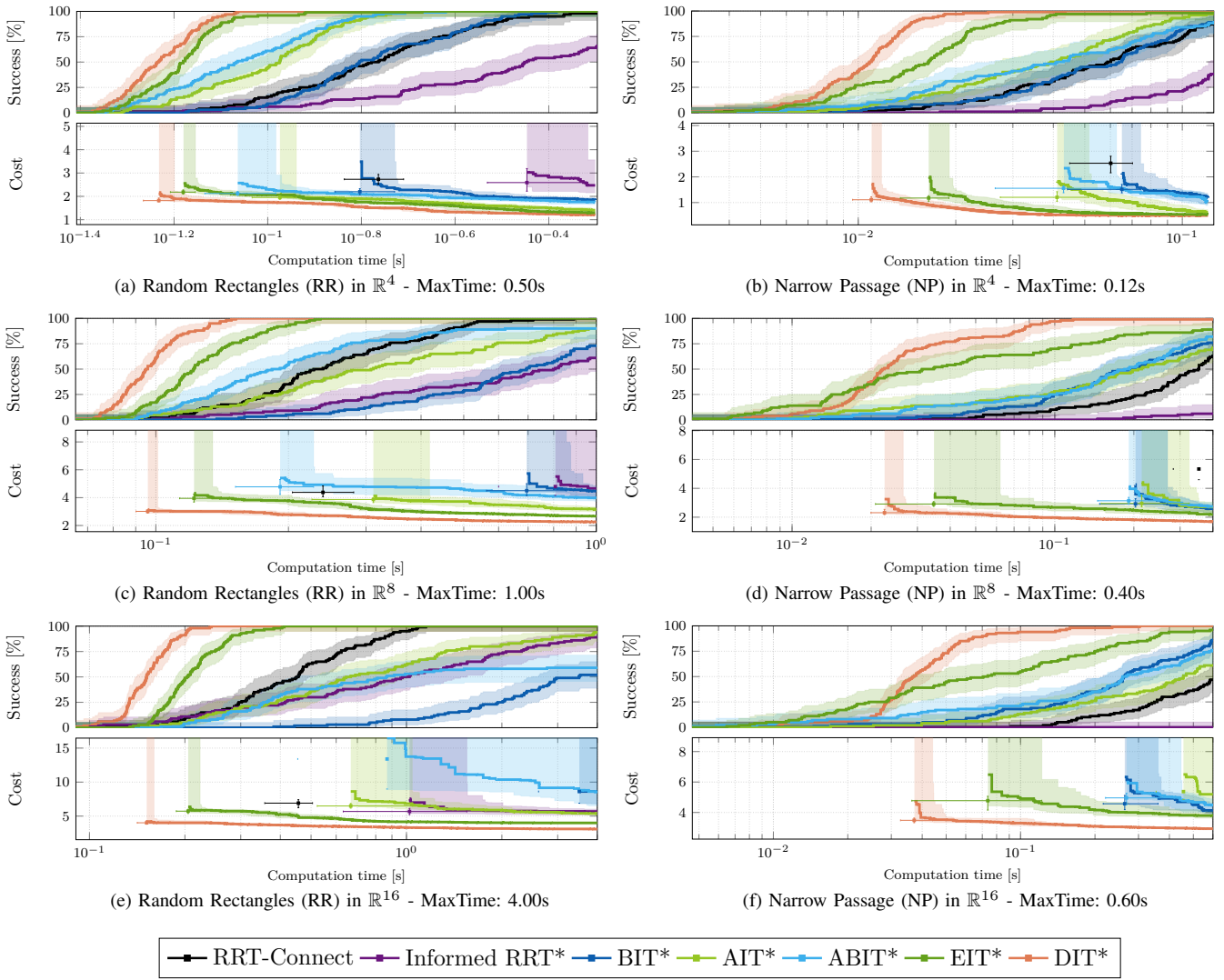
\begin{figure*}[t!]
    \centering
    \begin{tikzpicture}
    \node[inner sep=0pt] (russell) at (4.1,8)
    {\includegraphics[width=0.49\textwidth]{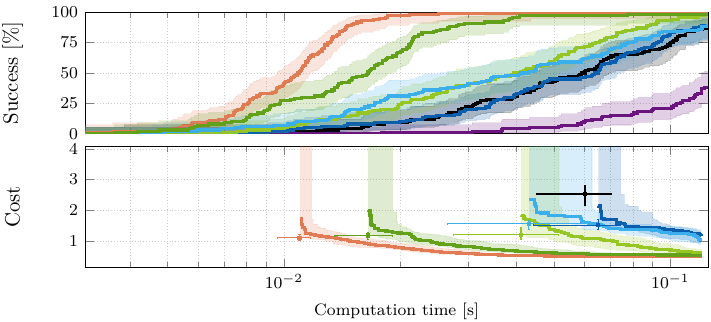}};
    \node[inner sep=0pt] (russell) at (4.1,3.5)
    {\includegraphics[width=0.49\textwidth]{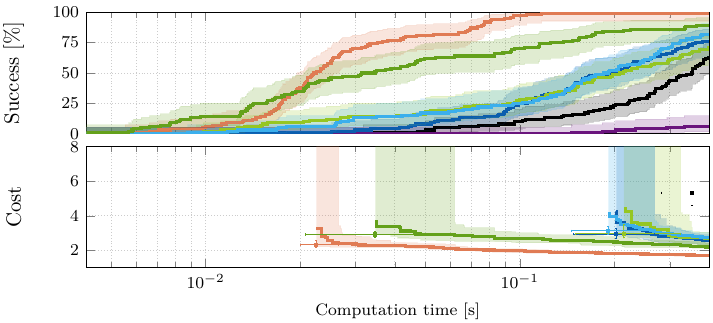}};
    \node[inner sep=0pt] (russell) at (4.1,-1)
    {\includegraphics[width=0.49\textwidth]{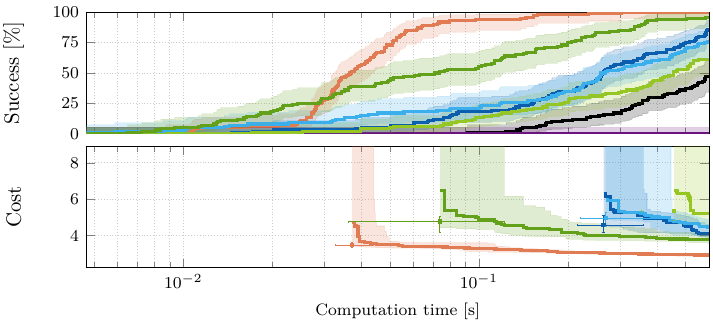}};

    \node[inner sep=0pt] (russell) at (-4.9,8)
    {\includegraphics[width=0.49\textwidth]{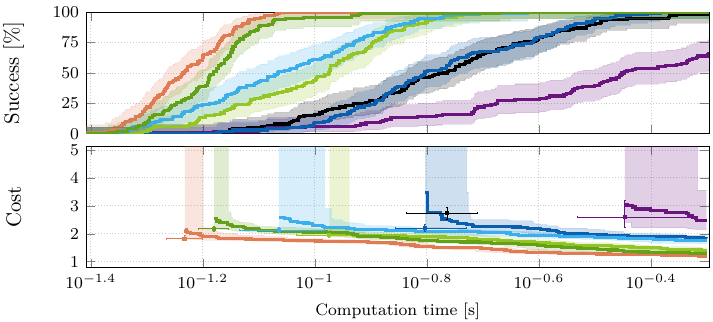}};
    \node[inner sep=0pt] (russell) at (-4.85,3.5)
    {\includegraphics[width=0.496\textwidth]{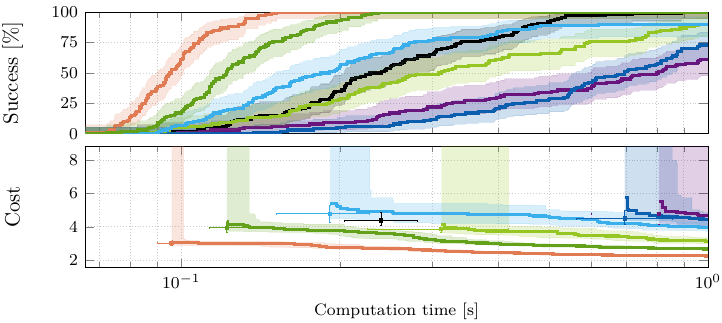}};  
    \node[inner sep=0pt] (russell) at (-4.9,-1){\includegraphics[width=0.49\textwidth]{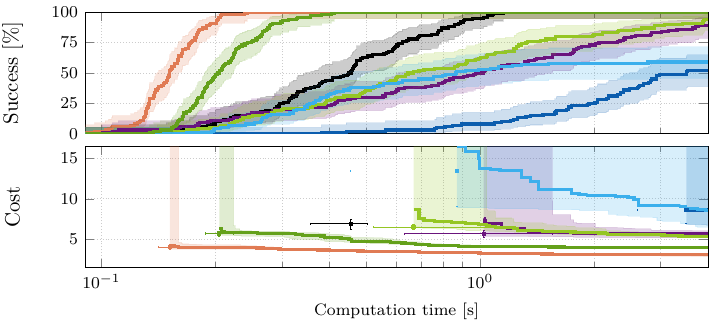}};

    \node[inner sep=0pt] (russell) at (0.0,-4.05){\includegraphics[width=0.8\textwidth]{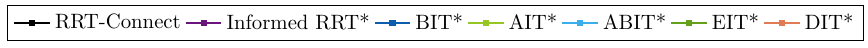}};

    \node at (-4.5,5.8) {\footnotesize (a) Random Rectangles (RR) in $\mathbb{R}^4$ - MaxTime: 0.50s};
    \node at (-4.5,1.3) {\footnotesize(c) Random Rectangles (RR) in $\mathbb{R}^8$ - MaxTime: 1.00s};
    \node at (-4.5,-3.2) {\footnotesize(e) Random Rectangles (RR) in $\mathbb{R}^{16}$ - MaxTime: 4.00s};

    \node at (4.5,5.8) {\footnotesize (b) Narrow Passage (NP) in $\mathbb{R}^4$ - MaxTime: 0.12s};
    \node at (4.5,1.3) {\footnotesize(d) Narrow Passage (NP) in $\mathbb{R}^8$ - MaxTime: 0.40s};
    \node at (4.5,-3.2) {\footnotesize(f) Narrow Passage (NP) in $\mathbb{R}^{16}$ - MaxTime: 0.60s};

    \end{tikzpicture}
    \vspace{-1.5em} 
    \caption{Detailed experimental results from Section~\ref{subsec:experi} are presented above. Fig. (a), (c), and (e) depict ten random rectangle outcomes in $\mathbb{R}^4$, $\mathbb{R}^8$, and $\mathbb{R}^{16}$, respectively. Panel (b) showcases narrow passage experiments in $\mathbb{R}^4$, while panels (d) and (f) demonstrate in $\mathbb{R}^8$ and $\mathbb{R}^{16}$. In the cost plots, boxes represent solution cost and time, with lines showing cost progression for an almost surely optimal planner (unsuccessful runs have infinite cost). Error bars provide nonparametric 99\% confidence intervals for solution cost and time.}
    \label{fig: result}
    \vspace{-1.8em}
\end{figure*}

We integrated the Direction Informed Trees (DIT*) algorithm into the Open Motion Planning Library (OMPL)~\cite{sucan2012open}, the Planner-Arena benchmark database~\cite{moll2015benchmarking}, and used Planner Developer Tools~\cite{gammell2022planner} for evaluation.
DIT* was tested against existing algorithms in both simulated random scenarios (Fig.~\ref{fig: testEnv}) and real-world manipulation problems (Fig.~\ref{fig: dual_arm_setup} and Fig.~\ref{fig:kitchenModel}). The comparison involved official versions of RRT-Connect, Informed RRT*, BIT*, AIT*, ABIT*, and EIT* sourced from the OMPL. The primary objective was the minimization of path length $c^\textit{med}_\textit{init}$ within a constrained time limit and to converge faster median initial solution finding time $t^\textit{med}_\textit{init}$. The RGG constant $\eta$ was consistently set to 1.001, and 
the rewire factor was uniformly set to 1.1 for all planners.

For RRT-based algorithms, a goal bias of 5\% was incorporated, with maximum edge lengths of 0.5, 1.25, and 3.0 in $\mathbb{R}^4$, $\mathbb{R}^8$, $\mathbb{R}^{16}$. Batch sampling planners maintained a fixed sampling of 100 states per batch across different dimensionalities of the state space. These planners employed the Euclidean distance and effort as a heuristic, respectively. 

DIT* utilizes a \textit{direction filter} with weight parameters $\omega_1 = 0.6, \omega_2 = 0.4$ to filter nearest neighbors bias goal and updates edge \textit{direction cost} heuristics with a tuned effort threshold $(\mu=5000)$ to enhance initial path convergence.
\subsection{Simulation Experimental Tasks}\label{subsec:experi}
The planners were subjected to testing across three distinct problem domains: $\mathbb{R}^4$, $\mathbb{R}^8$, and $\mathbb{R}^{16}$. 
In the first test environment, we introduced random widths to \textit{axis-aligned hyperrectangles} generated arbitrarily within the \textit{$\mathcal{C}$-space} (Fig. \ref{fig: testEnv}a). Distinct random rectangle (RR) problems were formulated for each dimension of the \textit{$\mathcal{C}$-space}, and we executed each planner 100 times for every instance. Fig.~\ref{fig: result}a, \ref{fig: result}c, and \ref{fig: result}e show that DIT* achieved the highest success rates.

In the second scenario, a constrained environment resembling a narrow passage (NP) with a slim gap (0.02) was simulated, allowing valid paths in only one general direction for non-intersecting solutions (Fig. \ref{fig: testEnv}b). Each planner underwent 100 runs, and the computation time for each benchmarked planner is demonstrated in the figure's label. The overall success rates and median path lengths for all planners are depicted in Fig.~\ref{fig: result}b, \ref{fig: result}d, and \ref{fig: result}f.
It shows that DIT* quickly finds
the initial solution across dimensions with minimal time, while EIT* takes longer to find the initial solution.

As shown in Table \ref{tab:benchmark}, DIT* consistently outperforms SOTA in initial median time ($t^\textit{med}_\textit{init}$) and initial path cost (path length, i.e., $c^\textit{med}_\textit{init}$) across various scenarios. In the $\text{RR}-\mathbb{R}^{8}$ scenario, DIT* achieves a $c^\textit{med}_\textit{init}$ of 3.0148, which is 23.68\% shorter than EIT*. Moreover, in higher dimension, the $\text{RR}-\mathbb{R}^{16}$ scenario, DIT* further reduces convergence time $t^\textit{med}_\textit{init}$ by 26.04\%.

In more confined settings, such as $\text{NP}-\mathbb{R}^{16}$, the level of improvement is even more substantial, with DIT* achieving a 45.08\% reduction in $t^\textit{med}_\textit{init}$. Overall, the table highlights the superior efficiency of DIT* in reducing initial median times. DIT* filters less favorable nearest neighbors, reducing the computational time. The optimal edges carry directional information, which biases exploration toward the goal.

\begin{table}[t]
\caption{Benchmarks evaluation comparison (Fig.~\ref{fig: result})}
\centering
\resizebox{0.485\textwidth}{!}{
\begin{tabular}{p{1.5cm}ccccccc}
 \hline
 & \multicolumn{3}{c}{${\text{Effort Informed Trees}}$} & \multicolumn{3}{c}{$\textcolor{purple}{\text{Direction Informed Trees}}$} &\multirow{2}*{\large$t^\textit{med}_\textit{init}\color{purple}\Uparrow$ (\%)}\\
 \cmidrule(lr){2-4} \cmidrule(lr){5-7}
    &$t^\textit{med}_\textit{init}$ &$c^\textit{med}_\textit{init}$ &$c^\textit{med}_\textit{final}$ &$t^\textit{med}_\textit{init}$ &$c^\textit{med}_\textit{init}$ &$c^\textit{med}_\textit{final}$ \\
 \toprule

    $\text{RR}-\mathbb{R}^4$   &0.0660   &2.1817   &1.2943 &0.0585 &\textcolor{purple}{1.8194} &1.2040 &11.36  \\
    $\text{RR}-\mathbb{R}^8$   &0.1221   &3.9503   &2.6709 &0.0958 &\textcolor{purple}{3.0148} &2.2551  &{21.54}  \\
    $\text{RR}-\mathbb{R}^{16}$   
    &{0.2047} &{5.7269} &{3.9990} &{0.1514} &\textcolor{purple}{4.0179} &{3.1000}  &\textcolor{purple}{26.04}\\
\midrule    
    $\text{NP}-\mathbb{R}^4$    &0.0165 &1.1758 &0.5373 &\textcolor{purple}{0.0110} &1.1095 &0.4944  &\textcolor{purple}{33.33}\\
    $\text{NP}-\mathbb{R}^8$  &0.0346  &2.9125  &2.1870 &\textcolor{purple}{0.0225} &2.3104 &\textcolor{purple}{1.6899} &\textcolor{purple}{34.97}  \\
    $\text{NP}-\mathbb{R}^{16}$   &0.0712   &4.7750   &3.7961 &\textcolor{purple}{0.0391} &\textcolor{purple}{3.4845} &\textcolor{purple}{2.9378}  &\textcolor{purple}{\textbf{45.08}}  \\
\bottomrule
\end{tabular}} \label{tab:benchmark}
\vspace{-1.7em} 
\end{table}
\subsection{Real-world Path Planning Tasks}\label{subsec:experiReal}
We compare DIT* with EIT*, AIT*, and BIT* across various settings to evaluate their performance with the 7-DoF base-manipulator robot (DARKO) and dual-arm manipulation (14-DoF) in converging to the optimal solution cost and success rate over 30 runs. The \textit{cable routing} - ENV (Fig.~\ref{fig: dual_arm_setup}) features navigating cable fixtures holder obstacles, \textit{kitchen} - ENV (Fig.~\ref{fig:kitchenModel}) environments involve navigating through cluttered, narrow spaces. Each scenario requires finding a collision-free path from the start state to the goal region.

In the kitchen model setup, all planners were given 3.0s to address the problem. Over 30 trials, DIT* achieved 100\% success with a median solution cost of 16.84. EIT* demonstrated a 93.33\% success rate but had a higher median solution cost of 22.71. AIT* achieved a success rate of 90.00\% with a median solution cost of 25.97, while BIT* achieved an 83.33\% success rate with a median solution cost of 29.57. Throughout the experiment, DIT* exhibited a more goal-oriented behavior and consistently achieved higher success rates with reduced solution costs.

In the dual-arm setup, all planners were allotted 5.0s to address the problem. Over 30 runs, DIT* achieved a success rate of 96.67\% with a median solution cost of 15.48. EIT* achieved a 90.00\% success rate but with a median solution cost of 21.85. AIT* achieved an 86.67\% success rate with a median solution cost of 25.26, while BIT* achieved a 73.33\% success rate with a median solution cost of 30.06. DIT* achieved the shortest path length for both arms.

In short, compared with BIT*, AIT*, and EIT*, our proposed DIT* achieves the best performance in finding the initial solution and converging to the optimal solution.

%% file: sections/sec6_conclusion.tex
\section{Conclusion}
This paper proposes Direction Informed Trees (DIT*), a sampling-based planner that integrates direction filter and direction cost heuristics methods. The direction filter prunes redundant states in the nearest neighbor, while the direction cost heuristics are quantified by weighted similarity indexes to guide search directions to optimize local exploration. Then, the probabilistic completeness and asymptotic optimality were guaranteed.
Through simulations across various dimensions and real-world experiments, we demonstrated that the DIT* achieves faster initial path convergence and shorter path lengths compared to the SOTA planner in a robust manner. Moreover, real-world tasks further demonstrate the effectiveness of our method in practical applications.



%% file: sections/sec7_acknowledgements.tex
\section{Acknowledgements}
\thanks{$^{\dagger}$The authors acknowledge the financial support by the Bavarian State Ministry for Economic Affairs, Regional Development and Energy (StMWi) for the Lighthouse Initiative KI.FABRIK (Phase 1: Infrastructure as well as the research and development program under grant no. DIK0249).}

\thanks{$^{\ddagger}$The authors acknowledge the financial support by the Federal Ministry of Education and Research of Germany (BMBF) in the programme of “Souverän. Digital. Vernetzt.” Joint project 6G-life, project identification number 16KISK002.}